\pdfoutput=1
\documentclass[11pt]{article}

\usepackage{acl}
\usepackage{times}
\usepackage{latexsym}
\usepackage{tipa}
\usepackage{amsmath}
\usepackage{amsfonts} 
\usepackage{booktabs}
\usepackage{multirow}
\usepackage{adjustbox}
\usepackage{longtable}
\usepackage[T1]{fontenc}
\usepackage[inline,shortlabels]{enumitem}
\usepackage[utf8]{inputenc}

\usepackage{microtype}
\usepackage{inconsolata}
\usepackage{titlesec}
\titlespacing*{\paragraph}{0pt}{4pt}{4pt}
\titlespacing*{\subsection}{0pt}{4pt}{2pt}
\title{The taste of IPA\beersEmoji: Towards open-vocabulary keyword spotting and forced alignment in any language}

\author{Jian Zhu\textsuperscript{\textipa{B},\textipa{\oe}} \hspace{0.2in} Changbing Yang\textsuperscript{\textipa{B},\textipa{\oe}} \hspace{0.2in} Farhan Samir\textsuperscript{\textipa{B},\textipa{\oe}} \hspace{0.2in} Jahurul Islam\textsuperscript{\textipa{B}}\\ 
\textsuperscript{\textbf{\textipa{B}}}Department of Linguistics, University of British Columbia \\
\textsuperscript{\textbf{\textipa{\oe}}} Natural Language Processing Group, University of British Columbia \\
  \texttt{jian.zhu@ubc.ca}\hspace{0.2in}\texttt{\{cyang33,fsamir\}@mail.ubc.ca} \\}

\newcommand\ourclap{\textsc{Clap-Ipa}}
\newcommand\ouraligner{\textsc{Ipa-Aligner}}
\newcommand\fleurs{\textsc{Fleurs-Ipa}}
\newcommand\mswc{\textsc{Mswc-Ipa}}
\newcommand\doreco{\textsc{Doreco-Ipa}}
\newcommand\ipapack{\textsc{IpaPack}}
\newcommand{\beersEmoji}{\includegraphics[height=1.5em,trim=0 .4em 0 0]{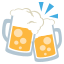}}

\begin{document}
\maketitle
\begin{abstract}
In this project, we demonstrate that phoneme-based models for speech processing can achieve strong crosslinguistic generalizability to unseen languages. We curated the \ipapack, a massively multilingual speech corpora with phonemic transcriptions, encompassing 115 languages from diverse language families, selectively checked by linguists. Based on the \ipapack{}, we propose \ourclap{}, a multilingual phoneme-speech contrastive embedding model capable of open-vocabulary matching between arbitrary speech signals and phonemic sequences. The proposed model was tested on 95 unseen languages, showing strong generalizability across languages. Temporal alignments between phonemes and speech signals also emerged from contrastive training, enabling zeroshot forced alignment in unseen languages. We further introduced a neural forced aligner \ouraligner{} by finetuning \ourclap{} with the Forward-Sum loss to learn better phone-to-audio alignment. Evaluation results suggest that \ouraligner{} can generalize to unseen languages without adaptation. 
\end{abstract}
\section{Introduction}
\label{sec:intro}

The diversity of human speech presents a formidable challenge to multilingual speech processing systems. Recently, accumulating evidence indicates that scaling up the multilingual data can tremendously improve the performance of multilingual speech processing \cite{conneau2020unsupervised,babu2021xls,radford2023robust,pratap2023scaling}. However, it remains incredibly difficult, if not impossible, to gather large-scale data from every language in the world. It is becoming increasingly critical to develop speech processing systems that generalize to arbitrary unseen languages. 

Despite the seeming diversity, sounds of human speech are highly constrained by the anatomical structure of the human vocal tract, which is universally shared by all humans \cite{gick2013articulatory}. Typological research has also shown that most, if not all, human speech can be represented by around 150 phonemes and diacritics \cite{moran2014phoible,gordon2016phonological}. The limited degrees of freedom in human articulation have enabled phoneticians and linguists to craft universal symbolic representations of human speech, that is, the \textbf{International Phonetic Alphabet (IPA)} \cite{international1999handbook}. 

Prior studies have shown that phoneme-based ASR models generalize to unseen languages \citep{li2020universal,xu22b_interspeech,glocker23_interspeech}. In this project, we aim to provide yet another positive answer to this central question: \textbf{can we build multilingual speech processing systems that generalize to arbitrary languages through the use of universal IPA symbols?} Specifically, we focus on two classic tasks in speech processing, \textbf{key word spotting (KWS)} and \textbf{forced alignment}. KWS is a task of identifying specific keywords in streaming speech, whereas forced alignment refers to aligning intervals of a speech signal to a given sequence of phonetic symbols. Both tasks are relevant in many practical applications such as voice assistant, speech synthesis, language documentation, etc. Yet neither task has been tackled with general systems that generalize to all languages. 

%Key Word Spotting (KWS) is a classic task of identifying specific key words in streaming speech, with primary applications in voice assistants and interfaces. Traditional KWS systems used to limit to a fixed set of key words in a certain language. It still remains a challenging task to develop open-vocabulary KWS systems that can deal with any input, and even more challenging to develop such a system for any language in the world \cite{lopez2021deep}. Currently most open-vocabulary KWS models still require a few samples to adapt to new words and struggles to perform zero-shot adaption. Effort has been made to multilingual KWS models but they are mostly limited to a dozen of languages \cite{mazumder21_interspeech,lei2023multilingual,reuter2023multilingual} and has not been shown to generalize to arbitrary unseen languages effectively.

 %However, multilingual KWS has yet to benefit from the scale of multilingual data. 

This study represents an attempt to build cross-linguistically generalizable systems for KWS and forced alignment. First, we present the \ipapack, a multilingual speech corpora in 115 languages with phonemic transcriptions, totaling over 1000 hours and carefully checked by trained linguists. Secondly, with the \ipapack, we
proposed Contrastive Language-Audio Pretraining with International Phonetic Alphabet (\ourclap{}), a phoneme-to-speech retrieval model with contrastive pretraining on phoneme-speech pairs. Evaluations on 95 unseen languages suggest that \ourclap{} is capable of performing zero-shot open-vocabulary KWS in any language without adaption, including languages not seen during training. 

Thirdly, we also introduce a multilingual forced alignment model, \ouraligner{}, that works for arbitrary languages. We noticed that alignments between phonemes and speech signals emerge from \ourclap{}, even with only sequence-level contrastive learning. Crosslinguistic zero-shot forced alignment can be achieved with \ourclap{}. After finetuning \ourclap{} with an alignment loss, we propose \ouraligner{} that can provide crosslinguistic word-level and phone-level alignment generalizable to unseen languages. 
Finally, our analysis indicates that phonemes, being shared across all languages, enhance knowledge transfer within training data, serving as more effective modeling units than texts in current multilingual tasks. 

We envision that our dataset and models will benefit more downstream tasks and applications in multilingual speech processing. To facilitate future research, we will release our dataset, scripts, and pre-trained models at: \url{https://github.com/lingjzhu/clap-ipa}. 

\section{Backgrounds}
\label{sec:related}

\subsection{Spoken keyword detection and retrieval}
%KWS is a classic task in spoken language processing, receiving wide attention from the speech community \cite{lopez2021deep}. %Deep neural networks have become the dominant paradigm in KWS systems since the past decade \cite[e.g.,][]{chen2014small,tang2018deep,rybakov20_interspeech,berg21_interspeech}.  %However, most of these systems are based on multiclass-classifications of a set of pre-defined keywords, making them hard to adapt to new keywords outside training set. 
Most research in keyword spotting focuses predominantly on English \cite[e.g.,][]{chen2014small,tang2018deep,rybakov20_interspeech,berg21_interspeech}. In recent years, there has been increased interest in building multilingual keyword detection systems that can adapt to new words or new languages through few-shot learning \cite{mazumder21_interspeech,lei2023multilingual,reuter2023multilingual}. While texts are the primary modeling units in most systems, studies are showing the effectiveness of using IPA symbols to achieve open-vocabulary generalization \cite{tanaka2001speech,shin22_interspeech,lee23d_interspeech,reuter2023multilingual}. 

Another approach for keyword matching is based on contrastive learning frameworks, notably CLAP \cite{wu2023large} and the subsequent CLARA \cite{noriy2023clara}. Contrastive learning also enables keyword retrieval systems based on semantics rather than the surface acoustic form \cite{duquenne2021multimodal,khurana2022samu,zhu-etal-2022-bootstrapping}. The contrastive learning paradigm has also been applied successfully to build open-vocabulary KWS systems \cite{nishu2023flexible}.

Nevertheless, existing multilingual KWS systems face limitations in terms of limited supported languages, and cannot achieve zero-shot adaptation. Built on these prior efforts, we scaled up the phoneme-based open-vocabulary KWS models to more languages to achieve crosslinguistic generalization.

\subsection{Forced alignment}
Forced alignment is another classic task in speech processing for segmenting speech into utterances, words, or phonemes. It is widely used for downstream tasks where phone or word durations are needed, including speech synthesis, speech assessment, language documentation, and speech corpora construction. Currently, some of the most popular forced alignment systems are still based on Hidden Markov Models (HMM), including the Montreal Forced Aligner (MFA) \cite{mcauliffe2017montreal}, WebMAUS \cite{kisler2012signal} and Forced Alignment and Vowel Extraction (FAVE) \cite{rosenfelder2011fave}. Recently, since neural networks gradually dominate speech processing, research in performing forced alignment with deep neural models is also gaining momentum \cite{kelley18_interspeech,kurzinger2020ctc,schulze2020joint,teytaut2021phoneme,teytaut2022study,zhu2022phone}. Neural models usually exhibit stronger performance over HMM-based systems. However, forced alignment systems are mostly set up to work in monolingual settings. Scant attention has been paid to the building of multilingual forced alignment systems that can work for multilingual languages simultaneously.

\begin{table*}[]
\begin{adjustbox}{width=0.99\textwidth,center}
\begin{tabular}{lccccccc}
\toprule
 & Train (hrs) & Dev (hrs) & Test (hrs) & Total (hrs) & Languages & Avg. Dur (hrs) \\\midrule
 VoxCommunis \citep{ahn-chodroff-2022-voxcommunis-2} &803.84 & -& -& 803.84& 38& 21.15\\\midrule
\ipapack\\\midrule
\hspace{0.5in}\fleurs{}&    544.02   & 73.46    &  162.06    & 779.54               &    77      &   10.12       \\
\hspace{0.5in}\mswc{} &    485.35    &  64.08   &  64.11    &     613.44           &   36       &   17.04         \\
\hspace{0.5in}\doreco{} &   13.70   &  -   &  5.29    &                18.99 & 44           &  0.44   \\\bottomrule     
\end{tabular}
\end{adjustbox}
\caption{Descriptive statistics of the \ipapack{} and a selected subset of VoxCommunis \cite{ahn-chodroff-2022-voxcommunis-2}.}
\vspace{-0.2in}
\label{tab:stats}
\end{table*}

\section{Dataset curation}
\label{sec:data}
Most speech corpora are distributed as audio-text pairs. In comparison, phonemically transcribed speech corpora are rare. Unlike text transcription, transcribing speech signals into IPA, or phonemic transcriptions often require years of expertise in phonetics, making it hard to create high-quality phonemic datasets at scale. However, these IPA symbols provide a universal representation of speech sounds such that any language can be transcribed symbolically. So IPA symbols can be used as a proxy to train multilingual speech processing systems. 
As a first step, we created large-scale phonemic transcriptions for public speech corpora, encompassing 115 languages across language families.  The transcription can be automated through \textbf{grapheme-to-phoneme conversion (G2P)}, a process of converting orthographic transcriptions into phonemic transcriptions through pronunciation dictionaries and/or statistical models.

\subsection{Phonemic transcriptions} We primarily made use of three existing multilingual speech datasets, FLEURS \cite{conneau2023FLEURS}, Multilingual Spoken Words Corpus (MSWC) \cite{mazumder2021multilingual} and DoReCo \cite{paschen2020building}. 

\paragraph{FLEURS}  We used two multilingual G2P systems, Epitran \cite{mortensen2018epitran} and CharsiuG2P \cite{zhu22_interspeech}, to create phonemic transcriptions. As these two systems cover an overlapping but slightly different set of languages, combining them allowed us to maximize the diversity of languages. Before preprocessing, we removed any texts with Arabic numbers or code-switching, as G2P systems cannot process them correctly.

Yet some Asian languages do not explicitly mark word boundaries with spaces. For Mandarin Chinese, G2PW \cite{chen22d_interspeech} was used to create the Pinyin romanizations, which were then mapped to IPA symbols. For Thai, we used PyThaiNLP \cite{pythainlp} to perform word segmentation and G2P. For Japanese, the word segmentation was first performed with Fugashi \cite{mccann-2020-fugashi} before G2P was applied. %We had to exclude Khmer data due to the failure to find a good tool for Khmer word segmentation.

\paragraph{MSWC} As MSWC is a word-level speech corpus, creating phonemic transcriptions was straightforward. CharsiuG2P and Epitran were deployed to transcribe the orthographic words to phonemic sequences. To strike a balance between diversity and quantity, we limited the maximum frequency to 50 to prevent high-frequency words from dominating the dataset. For words with more than 50 samples, only 50 of them will be randomly selected from the pool. After filtering, we ended up with 2.3 million spoken words, amounting to around 613 hours.

\paragraph{DoReCo} The original DoReCo data were distributed as hour-long recordings, so we segmented them into individual utterances based on the sentence boundaries in the provided annotations. For DoReCo, all languages were transcribed as phonemes using X-SAMPA \cite{wells1995computer} notations. We simply converted the X-SAMPA transcription to IPA symbols, as there is a one-to-one mapping between these two systems. Utterances with incomplete transcriptions or loud background noises were discarded. 

\subsection{Dataset validation} %Automatically generated multilingual datasets are often subject to various quality issues \cite{kreutzer-etal-2022-quality}. 
As G2P systems are based on rules or pronunciation dictionaries, they reflect how a word \textbf{should} be pronounced rather than how a word \textbf{is} pronounced. 
Given the high variability (e.g., phonetic reduction, coarticulation) in speech signals, it is not always possible for the G2P phonemic transcriptions to exactly match the audio.  We were aware that a true transcription does not always exist for every utterance \cite{ladefoged1988some,ladefoged1990revised}. Even trained phoneticians often disagree on the phonemic transcriptions of the same utterance, due to factors including psycho-acoustic constraints, phonetic training, and their mother tongue \cite{pitt2005buckeye,Heselwood2013}. 

Two authors (trained phoneticians) listened to at least ten random samples in each language to determine the transcription quality. We applied a relatively relaxed standard for the generated transcriptions: as long as the speech signal approximately matches more than 80\% of the transcription, it is considered valid. While we made our best efforts to validate the transcription quality, we acknowledge that there are still transcription errors in the dataset. 
A summary of the \ipapack{} is presented in Table~\ref{tab:stats}. To augment our current dataset, we also included a filtered subset of VoxCommuis Corpus \cite{ahn-chodroff-2022-voxcommunis-2}, which is a multilingual speech corpora created in a similar workflow, though with slightly different pronunciation dictionaries and G2P tools. Detailed information on individual languages of the VoxCommuis Corpus is at Appendix~\ref{app:dataset_stats}

\section{Method}

\subsection{Contrastive learning for KWS}
%Given a speech signal $\mathbf{S}=[s_1,s_2,\cdots,s_m]$ and a phonemic sequence $\mathbf{P}=[p_1,p_2,\cdots,p_n]$, $n\leq m$, our task of phoneme-to-speech retrieval is learn two encoders $f_S$ and $f_T$ to map both sequences into fixed-dimensional vectors in the same latent space, such that it can rank samples based on the similarity function $sim(f_S(\mathbf{P}),f_T(\mathbf{S}))$, where $sim(\cdot)$ is usually the cosine similarity $sim(\mathbf{x},\mathbf{y}) = \frac{\mathbf{xy}}{|\mathbf{x}||\mathbf{y}|}$. 

Here we adopt the same contrastive learning framework as CLAP \cite{wu2023large}, as it has been proven to be one of the most effective strategies for learning high-quality cross-modal representations. There are two separate encoders to process phoneme sequence $\mathbf{P}\in\mathbb{R}^{N\times 1}$ and speech MFCC features $\mathbf{S}\in\mathbb{R}^{T\times K}$, transforming them into phoneme embedding and speech embedding. %The CLIP loss function \cite{radford2021learning} relies on the softmax-based InfoNCE loss \cite{oord2018representation} to contrast a data sample with its positive and negative samples, which has been widely adopted for multi-modal or uni-modal contrastive training. 
In this study, we use the \textbf{SigLIP loss}, a simpler sigmoid-based loss that is shown to be as effective as the softmax-based CLIP loss \cite{zhai2023sigmoid}. Given two normalized embeddings $\boldsymbol{x}_i\in\mathbb{R}^D = f_S(\mathbf{P}_i)$ and $\boldsymbol{y}_i\in\mathbb{R}^D = f_T(\mathbf{S}_i)$, it is defined as follows.
\begin{equation}
 \mathcal{L} = -\frac 1 {|\mathcal{B}|} \sum_{i=1}^{|\mathcal{B}|} \sum_{j=1}^{|\mathcal{B}|} \underbrace{\log\frac 1 {1+e^{z_{ij}(-t\mathbf{x}_i \cdot \mathbf{y}_j  + b)}}}_{\mathcal{L}_{ij}}
\end{equation}

where $t$ and $b$ are learnable parameters that were updated during training. $z_{ij}$ is the ground truth label, $z_{ij}=1$ for positive pairs and $z_{ij}=-1$ for negative pairs. Following the recommendation by \citet{zhai2023sigmoid}, we initialized  $t = \log 10$ and $b=-10$.

\paragraph{Speech encoder} The speech encoder has the same transformer encoder architecture as the Whisper's encoder. The weights were initialized with Whisper's pre-trained encoder weights, whereas the decoder was discarded. The original Whisper encoder does not accept attention masks, but these padding tokens can bias the model during pooling. So attention masks were also passed to the speech encoder and the final fixed-dimensional embedding through mean pooling on non-padded hidden states. %Self-attention pooling has been shown to outperform mean and max pooling for speech embeddings. 
For speech data augmentation, SpecAugment \cite{park2019specaugment} was applied during training using the default hyperparameters in Whisper training.

\paragraph{Phoneme tokenizer} We trained a specialized tokenizer to encode all base IPA symbols and diacritics, including tonal notations, stress marks, and tie bars for affricates. Upon inspection, we noticed that the IPA transcriptions were inconsistent across languages. For example, tie bars were inconsistently labeled (e.g., [t\textipa{S}] vs. [\t t\textipa{S}]) and stress marks tend to be a language-specific phenomenon \cite{gordon2017acoustic}. Yet we did not perform normalization on these idiosyncratic labels to preserve the diversity of our data. %so that the final phoneme encoder can accept the highly variable inputs in practice. All other non-IPA characters were removed from our data. 
The phoneme tokenizer was trained using the unigram algorithm \cite{kudo2018subword} with \texttt{sentencepiece} package \footnote{\url{https://github.com/google/sentencepiece}}. The tokenizer was trained on all phonemic transcriptions in our datasets, with a vocabulary of 450 and byte-fallback for unknown characters. 

\paragraph{Phoneme encoder}  For the phoneme encoder, we used the BERT architecture \cite{devlin-etal-2019-bert} with mean pooling of the final hidden states as the fixed-dimensional representation. The phoneme encoder was pre-trained on a corpus of phonemic transcriptions using standard masked language modeling (MLM) as detailed in \cite{devlin-etal-2019-bert}. Given that phoneme sequences are of less complexity than texts, the masking probability was set to 30\%. The training data were pooled from diverse sources, including the \ipapack{}, pronunciation dictionaries in CharsiuG2P \cite{zhu22_interspeech}, and Vox Communis \cite{ahn-chodroff-2022-voxcommunis-2}. The final pretraining corpus consists of 11 million samples in more than 110 languages. We pre-trained three phoneme encoders of different sizes, matching hyperparameters including the number of layers, hidden dimensions, and the number of attention heads to the corresponding Whisper encoder (\texttt{tiny}, \texttt{base} and \texttt{small}).

\subsection{Forced alignment}
We noticed that phoneme-to-speech alignment emerged from \ourclap{} on the pairwise cosine similarity matrix computed with the token-wise hidden states of phone and speech encoders. We introduce a simple algorithm to derive the alignment between phonetic units and speech signals, with control over the temporal resolution of speech frames and the granularity of phonetic sequences.

\begin{figure}
    \centering
    \adjustbox{width=\linewidth,trim={4.5cm 3.5cm 4.5cm 3cm},clip}{
    \includegraphics{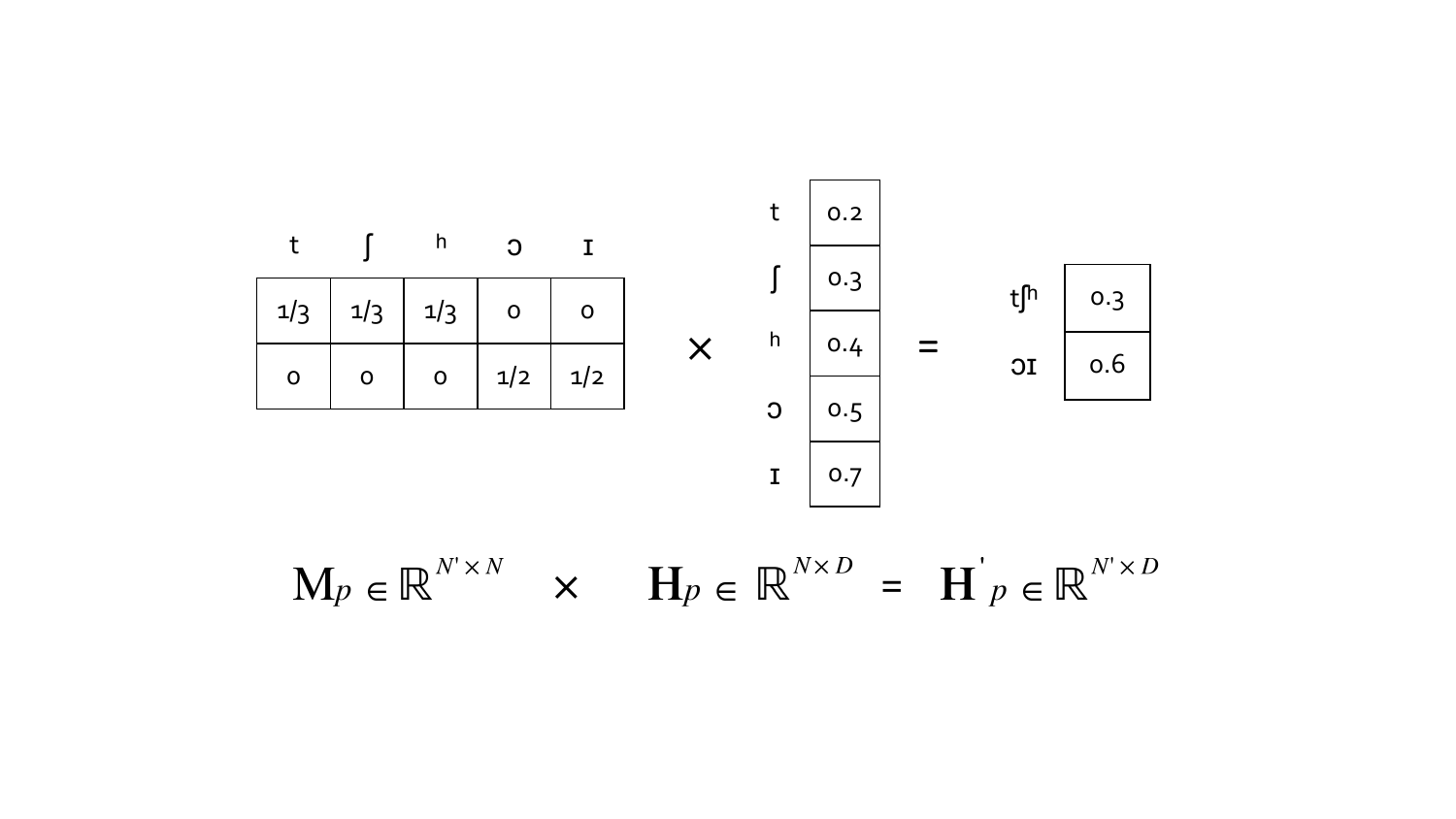}
    }
    \caption{Illustration of adaptive average-pooling of phoneme representations, $\mathbf{M_p}\mathbf{H_p} = \mathbf{H_p^{\prime}}$.}
    \vspace{-0.25in}
    \label{fig:pooling}
\end{figure}

\paragraph{Adaptive average pooling} While we expect the forced aligned units to be natural phonetic units like phonemes and words, due to tokenization, the hidden states of phone encoders correspond to a character or byte unit rather than a natural phonetic unit. A sequence of phonemes or words of length $N^\prime$ might be tokenized into a character or byte sequence of length $N, N\geq N^{\prime}$.  %A phoneme could consist of multiple IPA units through the combination of phone units and diacritics. To deal with the diversity of potential combinations, we tokenized phonemes into individual characters or bytes. 
We define an adaptive average-pooling mask $\mathbf{M}_p\in\mathbb{R}^{N^{\prime}\times N}$ to downsample the hidden representations. Through this pooling mask, consecutive hidden states belonging to one phoneme or one word were averaged to one fixed dimensional vector, such that each output hidden state after pooling corresponds to a natural phonetic unit (see Fig~\ref{fig:pooling}). This ensures that our forced alignment algorithm works for any level of phonetic units. 

We can also define a similar adaptive average-pooling mask for speech representations $\mathbf{M}_s\in\mathbb{R}^{T^{\prime}\times T}$ to downsample them from length $T$ to $T^\prime$. For word-level alignments that don't require high temporal resolution, we can compress the length of the speech hidden states by controlling the pooling window length and frameshift. 

\paragraph{Zeroshot forced alignment} Given two sequences of hidden states $\mathbf{H}_s\in\mathbb{R}^{T\times D}$ and $\mathbf{H}_p\in\mathbb{R}^{N\times D}$ produced by the speech encoder and phone encoders, adaptive average-pooling masks $\mathbf{M}_p\in\mathbb{R}^{N^{\prime}\times N}$ and $\mathbf{M}_s\in\mathbb{R}^{T^{\prime}\times T}$ are used to transform them into more compact representations $\mathbf{H_s^{\prime}}\in\mathbb{R}^{T^{\prime}\times D}$ and $\mathbf{H_p^{\prime}}\in\mathbb{R}^{N^{\prime}\times D}$.

\[
\mathbf{H_s^{\prime}} = \text{Normalize}(\mathbf{M_s}\mathbf{H_s}\text{,dim=-1})
\]
\[
\mathbf{H_p^{\prime}} = \text{Normalize}(\mathbf{M_p}\mathbf{H_p}\text{,dim=-1})
\]
\[
\mathbf{D} = \mathbf{H_s^{\prime}}\mathbf{H_p^{\prime\top}}/\tau
\]
where $\tau$ is the fixed temperature parameter and was set to 0.05 by default. The pairwise similarity matrix $\mathbf{D}\in\mathbb{R}^{T^{\prime}\times N^{\prime}}$ is used to derive the temporal monotonic alignment between phonetic units and speech frames through \textbf{dynamic time warping (DTW)}, even if \ourclap{} had never between trained on alignment labels. 
 
\paragraph{Finetuning} To further enhance the performance of forced alignment, we introduce \ouraligner{} by finetuning \ourclap{} with the \textbf{Forward-Sum Loss}, which has been shown to be effective in learning monotonic alignments between speech and phonemes \cite{shih2021rad,badlani2022one,zhu2022phone}. 

\[
\mathcal{L} = \mathcal{L}_{ForwardSum}(\mathbf{D})
\]
This alignment learning loss function relies on the forward-sum algorithm in classic HMMs to maximize the likelihood of text sequence given speech sequences, while enforcing the monotonic constraint of alignment (see \citet{shih2021rad} for detailed derivations). The Forward-Sum loss requires a good prior alignment to converge to meaningful results, so we did not report failure results from randomly initialized models.

During finetuning, we only average-pooled the phoneme representations at the phoneme and kept the original speech representations (by setting $\mathbf{M}_s$ to the identity matrix $\mathbf{I}$). In inference, for phoneme alignment, we pooled the phoneme representations at the phoneme-level and kept the original speech representations. For word alignment, the phoneme representations were pooled at the word-level and the speech representations were average-pooled with a window length of 3 and a frameshift of 2.

\begin{table*}[!th]
\centering
\footnotesize
\begin{tabular}{@{\extracolsep{4pt}}ccccc@{}}
\toprule
\multirow{2}{*}{Method} & \multicolumn{2}{c}{LibriPhrase-Easy} & \multicolumn{2}{c}{LibriPhrase-Hard}  \\\cline{2-3} \cline{4-5}   
                       & EER(\%) $\downarrow$   & AUC(\%) $\uparrow$    & EER(\%) $\downarrow$   & AUC(\%) $\uparrow$         \\\midrule
CMCD \cite{shin22_interspeech}    &   8.42      &      96.7        &  32.90          &     73.58           \\
PhonMatchNet \cite{lee23d_interspeech}    &   2.80         &     99.29        &  18.82          &     88.52        \\
CED \cite{nishu2023flexible}     &   1.7         &    99.84      &  \textbf{14.4}          &     \textbf{92.7}   \\\midrule

\ourclap{}-\textsc{Text}  & 6.0 & 98.31 & 31.14 & 74.8 \\
\ourclap{}-\textsc{Phone}  & 1.3 & 99.88  & 23.03 & 84.58\\\midrule
\ourclap{}-\textsc{Fleurs} & 0.95 & 99.94  & 22.98 & 84.82 \\
\ourclap{}-\textsc{Vc} & 0.81 & 99.55 & 21.55 & 85.91 \\\midrule
\ourclap{}-tiny  & 0.68 & 99.96 & 20.85 & 86.58 \\
\ourclap{}-base  & 0.63 & 99.97 & 20.04 & 88.25 \\
\ourclap{}-small  & \textbf{0.56} & \textbf{99.97} & 18.62 & 88.82  \\
   \bottomrule      
\end{tabular}
\caption{Evaluation results on the English-only Libriphrase.}
\label{tab:libriphrase}
\end{table*}

\begin{table*}[!th]
\centering
\footnotesize
\begin{tabular}{@{\extracolsep{4pt}}ccccccccc}
\toprule
\multirow{2}{*}{Model}  & \multicolumn{2}{c}{\mswc{}} & \multicolumn{2}{c}{\fleurs{}} & \multicolumn{2}{c}{\textsc{UclaPhoneticCorpus}} &\multicolumn{2}{c}{\doreco{}}  \\\cline{2-3}\cline{4-5}\cline{6-7}\cline{8-9}
& Hit@1 $\uparrow$           & mAP $\uparrow$           & Hit@1 $\uparrow$           & mAP $\uparrow$ 
& Hit@1 $\uparrow$           & mAP $\uparrow$ 
& Hit@1 $\uparrow$           & mAP $\uparrow$ \\\midrule
\ourclap{}-\textsc{Text}           &     13.51             &      12.7          &      8.48               &   10.22  & - & - & - & -        \\
\ourclap{}-\textsc{Phone}           &   79.28              &   68.74               &   86.4               &   87.4 & - & - & - & - \\\midrule
\ourclap{}-\textsc{Fleurs} & 83.48 & 77.16 & 98.59 & 98.53 & 51.57 & 62.53 & 73.91 & 79.32 \\
\ourclap{}-\textsc{Vc} & 84.38 & 75.64 & 63.52 & 63.21 & 50.05 & 61.41 & 90.56 & 93.01\\\midrule
  \ourclap{}-tiny       &     82.58             &   76.32               &      98.85         &   98.86 & 51.71 & 62.62  & 95.46 & 96.84    \\
  \ourclap{}-base               &  \textbf{82.60}              &     \textbf{77.31}             &    \textbf{99.20}            &       \textbf{99.27}  & 52.17 & 63.90 & \textbf{96.54} & \textbf{97.77} \\
  \ourclap{}-small              &     81.98           &    75.35                &   97.61            &  97.98  & \textbf{55.05} & \textbf{65.93} & 91.46 & 94.41    \\                          
\bottomrule
\end{tabular}
\caption{Evaluation results on unseen languages.}
\vspace{-0.2in}
\label{tab:unseen_eval}
\end{table*}

\section{Experiments}
\subsection{Training details}
We trained three variants of models, \ourclap{}-tiny, \ourclap{}-base and \ourclap{}-small, all of them were matched to the default encoder parameters of Whisper \cite{radford2023robust}. The speech encoder and phoneme encoder were symmetric. Our training dataset included the training set of \ipapack{} plus the VoxCommunis speech corpora \cite{ahn-chodroff-2022-voxcommunis-2}. By default, all models were trained with paired speech recordings and their phonemic transcriptions.
For \ouraligner{}, we finetuned \ourclap{}-tiny, \ourclap{}-base and \ourclap{}-small on the same data excluding \mswc{}. All detailed hyperparameters can be found in Appendix~\ref{app:training}. %All models were trained for 100k iterations with the Adam optimizer \cite{KingBa15,loshchilov2018decoupled} with the same hypterparameters. 

For controlled comparison, we also trained two base models, \ourclap{}-\textsc{Text} and \ourclap{}-\textsc{Phone} on the same \fleurs{} and \mswc{} subset either with only phonemic or text transcriptions. These two models were matched in total parameters, training data, and all other hyperparameters during training. In another controlled experiment, we trained \ourclap{}-\textsc{Fleurs} and \ourclap{}-\textsc{Vc} either only on the \fleurs{} or the VoxCommunis, which would allow us to examine the impact of data size and language diversity. 

\subsection{Evaluation datasets} We evaluated the crosslinguistic generalizability of our models on several evaluation datasets covering a wide range of topologically diverse languages. Whenever possible, we made our best effort to include baseline models to contextualize our model performance. This was not always possible, because evaluating multilingual KWS and multilingual forced alignment on unseen languages are new tasks and in some cases we were not able to find open-source models for comparison. However, we hope that our models and results will become a baseline that spur more future research in this direction. 

\paragraph{Libriphrase} To compare with existing models, we first tested on a popular English KWS dataset, Libriphrase \cite{shin22_interspeech}, as an out-of-domain evaluation dataset, since our models were not trained on their training sets. We used \textbf{Equal Error Rate (EER)} and the \textbf{Area under Curve (AUC)} scores to compare model performance, consistent with prior studies. 

\paragraph{Unseen languages} We also evaluated all models on five unseen languages with typological diversity from \fleurs{} and \mswc{}. We isolated five language from \mswc{} and \fleurs{}, namely, Vietnamese (\texttt{vie}), Tamil (\texttt{tam}), Hausa (\texttt{hau}), Georgian (\texttt{geo}) and Odia (\texttt{ori}). For \fleurs{}, the test sets of these five languages were directly used. However, for \mswc{}, due to data scarcity, we pooled all training, validation, and tests of these five languages together to form a larger and more challenging benchmark. We further evaluated 95 (81 unseen) languages from the UCLA phonetic Corpus \cite{li2021multilingual} and 14 unseen languages from \doreco{}.  \textbf{Hit@1} and \textbf{Mean Average Precision (mAP)} were used to measure the cross-linguistic retrieval performance of all models. To avoid duplication, we only reported results on phoneme-to-speech retrieval, as the results of speech-to-phoneme and speech-to-speech retrieval were in the same range.

%To further validate the crosslinguistic generalizability of the proposed models, we tested them on 97 unseen languages from linguistic fieldwork recordings. We used the UCLA phonetic Corpus \cite{li2021multilingual} as a word-level test set and a subset of the DoReCo corpus \cite{paschen2020building} as an utterance-level test set. All together there were 97 unseen languages in these two test sets, which makes them suitable for evaluating the proposed model's capacity of crosslinguistic generalizability. 
\paragraph{Word and phoneme boundaries} To evaluate the performance of forced alignment, we made use of \textbf{F1} and \textbf{R-Value}, which were used in prior studies \cite{rasanen09b_interspeech,kreuk20_interspeech,zhu2022phone}.  If the predicted boundary is within the tolerance interval of the true boundary, it is considered a hit, otherwise a miss. Since each boundary marked the onset and the offset of consecutive phones, we only evaluated the phone onsets with a tolerance of 20ms and word onsets with a tolerance of 100ms. We used TIMIT \cite{garofolo1993darpa} as the English benchmark. \doreco{} also contains phoneme-level and word-level alignments, so we partitioned the \doreco{} into seen and unseen evaluation sets. Yet \ouraligner{} was never trained on any segmentation labels.

\begin{table}[!th]
\begin{adjustbox}{width=0.99\linewidth,center}
\begin{tabular}{@{\extracolsep{4pt}}ccccc@{}}
\toprule
\multirow{2}{*}{Method} & \multicolumn{2}{c}{TIMIT-Word} & \multicolumn{2}{c}{TIMIT-Phone}  \\\cline{2-3} \cline{4-5}   
                       & F1 $\uparrow$   & R-Val $\uparrow$   & F1 $\uparrow$   & R-Val $\uparrow$      \\\midrule
FAVE & - & - & 58.0 & 64.0 \\
MFA & - & - & 63.0 & 68.0 \\
Gentle &- & - & 48.0 & 56.0 \\
WebMAUS &- & - & \textbf{70.0} & \textbf{75.0} \\
W2V2-FC-20ms &- & - & 48.0 & 56.0 \\
W2V2-FS-20ms &- & - & 48.0 & 55.0 \\\midrule
\textsc{Zeroshot} \\\midrule                       
\ourclap{}-tiny  & 84.37 & 86.66 & 40.46 & 49.92 \\
\ourclap{}-base  & 78.61 & 81.73 & 36.16 & 46.59 \\
\ourclap{}-small  & 74.18 & 77.95 & 35.26 & 46.17  \\\midrule
\textsc{Finetuned} \\\midrule  
\ouraligner{}{}-tiny  & \textbf{86.84} & \textbf{88.75} & 57.31 & 63.66 \\
\ouraligner{}{}-base  & 86.55 & 88.51 & 60.86 & 66.67 \\
\ouraligner{}{}-small  & 82.33 & 84.76 & 52.54 & 59.53 \\
   \bottomrule      
\end{tabular}
\end{adjustbox}
\caption{Evaluation of forced alignment on \textsc{Timit}. Baseline results were retrieved from \citet{zhu2022phone}. The temporal resolution is 10ms for FAVE, MFA, Gentle, and WebMAUS and 20ms for the rest of the models.}
%\vspace{-0.3in}
\label{tab:timit}
\end{table}

%\vspace{-0.5in}

\begin{table*}[!th]
\centering
\footnotesize
\begin{tabular}{@{\extracolsep{4pt}}ccccccccc@{}}
\toprule
\multirow{2}{*}{Method} & \multicolumn{2}{c}{Seen-Word} & \multicolumn{2}{c}{
Seen-Phone}  & \multicolumn{2}{c}{Unseen-Word} & \multicolumn{2}{c}{Unseen-Phone} \\\cline{2-3} \cline{4-5} \cline{6-7} \cline{8-9}   
                       & F1 $\uparrow$   & R-Val $\uparrow$   & F1 $\uparrow$   & R-Val $\uparrow$  & F1 $\uparrow$   & R-Val $\uparrow$ & F1 $\uparrow$   & R-Val $\uparrow$    \\\midrule
\textsc{Zeroshot} \\\midrule                       
\ourclap{}-tiny  & 67.19 & 72.17 & 33.8 & 44.74 & 68.53  & 73.27 & 35.26 & 45.91 \\
\ourclap{}-base  & 57.19 & 63.9 & 29.09 & 41.30 &  58.35 & 64.91 & 32.03 & 42.59 \\
\ourclap{}-small  & 51.48 & 59.43 & 28.59 & 40.85  & 59.94 & 66.18 & 30.24 & 42.43 \\\midrule
\textsc{Finetuned} \\\midrule  
\ouraligner{}-tiny  & 74.18 & 77.99 & 47.08 & 54.93 & 76.33 & 79.82 & 48.96 & 56.55 \\
\ouraligner{}-base  & \textbf{78.30} & \textbf{91.47} & \textbf{48.04} & \textbf{55.70} & \textbf{80.71} & \textbf{83.52} & \textbf{50.32} & \textbf{57.67}\\
\ouraligner{}-small  & 72.24 & 76.37 & 44.89 & 52.97  & 73.63 & 77.52 & 46.46 & 54.38 \\
   \bottomrule      
\end{tabular}
\caption{Evaluation of forced alignment on \doreco{}. The word boundary metrics were calculated with 100ms tolerance, whereas the phone boundary was computed with 20ms tolerance.}
%\vspace{-0.25in}
\label{tab:doreco_boundary}
\end{table*}

\section{Results}
In this section, we summarize the main results for KWS and forced alignment. 

\paragraph{KWS} Evaluation results in 
Table~\ref{tab:libriphrase} suggests that \ourclap{} performs on par with the state-of-the-art models on LibrisPhrase-Easy, while not trained on the Libriphrase training set. Yet \ourclap{} failed to outperform state-of-the-art CED \citep{nishu2023flexible} in LibriPhrase-Hard, suggesting that language-specific finetuning is still necessary to maximize performance. Generally speaking, phoneme-based models are more effective than text-based models. 

For unseen languages, Table~\ref{tab:unseen_eval} indicates that phoneme-based models do generalize successfully to unseen languages across datasets. In contrast, the text-based model performs poorly in unseen languages, suggesting that orthographic texts are not very useful for crosslinguistic speech processing. Utterance-level retrieval appears to be much easier than word-level retrieval, a pattern quite consistent across datasets. Model size correlates with performance in seen languages but not with crosslinguistic generalizability. 

\paragraph{Forced alignment} While not trained on forced alignment explicitly, \ourclap{} shows some capabilities for crosslinguistic forced alignment even in zero-shot predictions on both seen and unseen languages (see Table~\ref{tab:timit} and Table~\ref{tab:doreco_boundary}). After finetuned with the ForwardSum loss, \ouraligner{} can perform competitively in English with some widely used HMM-based forced aligners, even though TIMIT was not part of its training dataset. For low-resource languages, \ouraligner{} also achieves good performance, regardless of whether the language has been seen during training or not. 

\begin{figure}
    \centering
    \includegraphics[width=\linewidth]{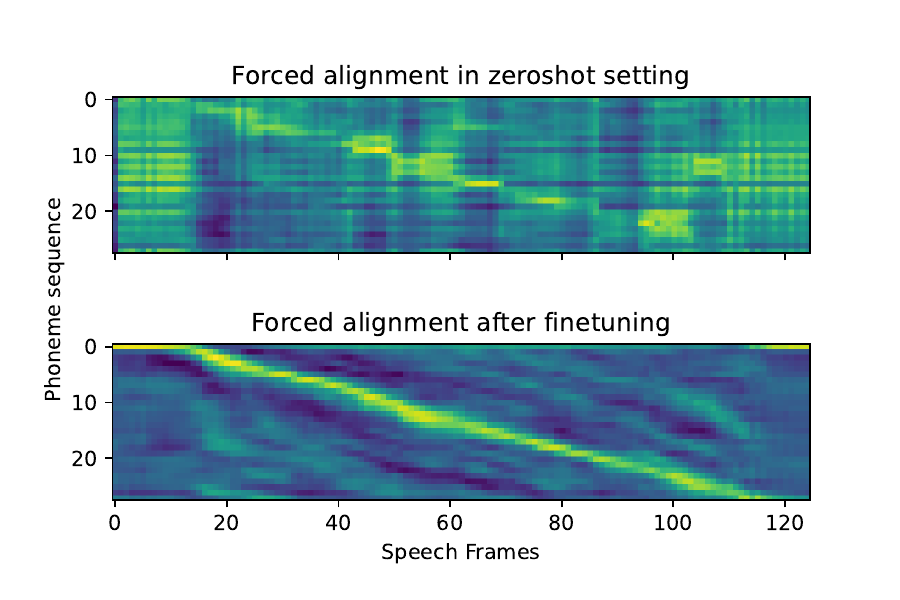}
    \caption{Illustration of forced alignment in an Evenki utterance. \ourclap{} exhibits vague monotonic alignment without finetuning (\textbf{Top}). After finetuning, \ouraligner{} learns salient monotonic alignment between speech and phonemes (\textbf{Bottom}).}
    %\vspace{-0.25in}
    \label{fig:alignment}
\end{figure}

\begin{table*}[!th]
    \centering
    %\small
    \begin{tabular}{c|c|l} \toprule
       Query  & Output Type & Retrieved candidates (ranked from high to low) \\\midrule
        \textipa{\'et\'a}  &  Most similar & \textipa{\`Et\'a}, \textipa{\'eta}, \textipa{l\=ata}, \textipa{\'et\'a}, \textipa{\`\i t\`a}, \textipa{ait:a}, \textipa{m\`Et\'a}, \textipa{me\|[ta}, \textipa{\t pt\'a}, aita, \textipa{\`Et\'E}, \textipa{at\super sa}, \textipa{\`at\'a}, \textipa{eit@} \\
            & Most dissimilar & \textipa{b\~O\.*ti}, \textipa{t\super ju:riS}, \textipa{sorNgi}, \textipa{aBu"Ru}, \textipa{mb\'uru\`u}, \textipa{sumbuN}, \textipa{buluz}, \textipa{S\`\i Z\`\i Z}, \textipa{tSungu} \\\bottomrule
    \end{tabular}
    \caption{Sample ranked phonemic sequences by \ourclap{}-small, given the speech query [\textipa{\'et\'a}].  }
    
    \label{tab:example}
\end{table*}

\begin{figure*}[!th]
    \centering
    \includegraphics[width=\textwidth]{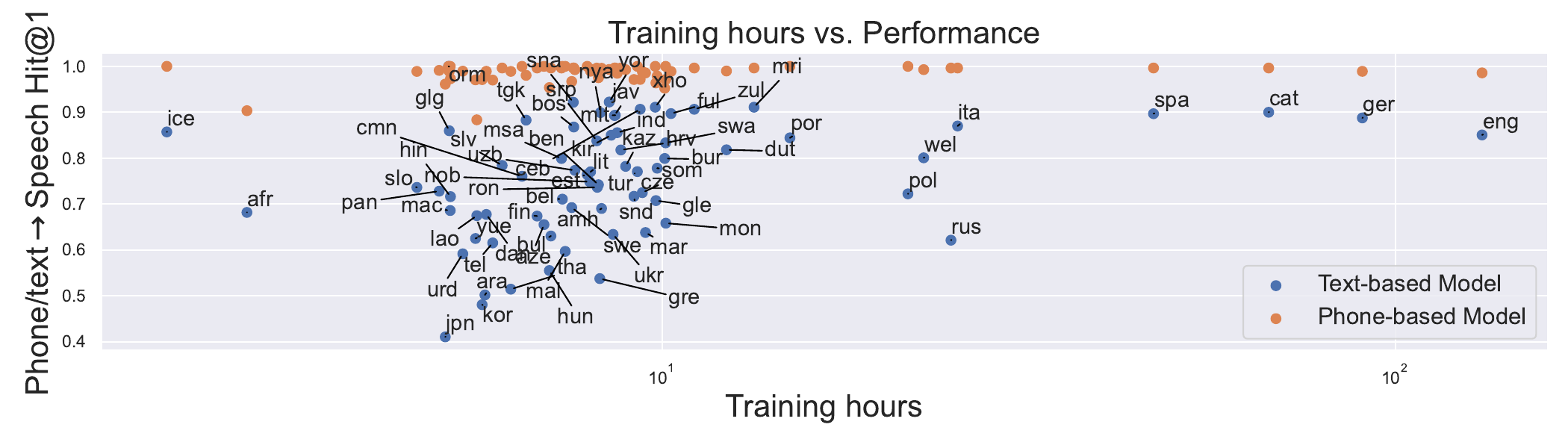}
    \caption{Correlation of model performance on individual languages with training hours by language. Languages are represented by their ISO 639-3 codes. While trained the exact same data, the phoneme-based model outperforms the text-based model in every single language, suggesting that phoneme-based modeling enables knowledge transfer across languages. }
    \label{fig:hours}
    %\vspace{-0.2in}
\end{figure*}

\section{Discussions}
In this section, we provide more in-depth answers to our research questions with the major findings of our experiments. 
\paragraph{Can phoneme-based models generalize cross-linguistically?} The evaluation results for \ourclap{} and \ouraligner{} in Table~\ref{tab:unseen_eval} and Table~\ref{tab:doreco_boundary} indicate that phoneme-based model exhibits strong generalization capabilities cross-linguistically in both KWS and forced alignment, even to unseen languages in zero-shot predictions. 

Generally speaking, all \ourclap{} models perform better on utterance-level datasets (\fleurs{} and \doreco{}) than on word-level datasets (\mswc{} and UCLA Phonetic Corpus), because the longer the phoneme sequence, the more likely that it is distinct in a pool of candidates.  For utterance-level datasets, \ourclap{} models achieve near-perfect scores on unseen languages (see Table~\ref{tab:unseen_eval}), indicating that phonemic representations do enable cross-linguistic generalization.

Table~\ref{tab:example} shows %the ranking of retrieved phonemic transcriptions (among 5000 potential candidates in 95 languages) with respect to the speech query [\textipa{\'et\'a}] in the UCLA Phonetic Corpus. 
that the similarity assigned by \ourclap{}-small was highly consistent with human perception. The top-ranked crosslinguistic candidates were extremely similar in articulatory features and syllable structure to the query. 

For forced alignment, even the zero-shot predictions using \ourclap{} can perform segmentation in unseen languages, especially at the word level. Interestingly, there were no significant differences between performance over seen and unseen languages. Though this result could be biased by the smaller number of unseen languages compared to seen languages (14 vs. 30), it still suggests that \ouraligner{} can perform crosslinguistic forced alignment without much adaptation. Finetuning the \ouraligner{} brings continued improvement over the zero-shot scenarios (see Fig~\ref{fig:alignment}).

\paragraph{Does the phoneme-based model generalize better cross-linguistically than the text-based model?} Text-based models are struggling to generalize to unseen languages, as these unseen languages have their distinct writing systems (e.g, Vietnamese and Tamil) that are not seen in training languages. Comparison between the text-based and phoneme-based models in Table~\ref{tab:libriphrase} and Table~\ref{tab:unseen_eval}
clearly shows that it is the use of phonemes as modeling units that brings strong crosslinguistic generalizability, since they can represent all languages using the same set of symbols.

\paragraph{Do the training hours of individual languages predict the performance of multilingual models?}  The number of training hours for individual languages does not predict the performance of language in phoneme-based models. All languages benefit from the multilingual knowledge transfer in phoneme-based modeling. 

It has been reported that there is a strong correlation between text-based multilingual ASR performance in individual languages and their training hours \cite{radford2023robust,rouditchenko23_interspeech}. We also confirm that, for text-based models, there is a moderate correlation between Hit@1 and the number of training hours (Spearman's $\rho: 0.42$; $p\leq 0.0002$). However, this correlation was not significant for the phoneme-based model (Spearman's $\rho: 0.14$; $p=0.22$). In Figure~\ref{fig:hours}, the phoneme-based model outperforms the text-based model in every language by a large margin, especially for languages with less training data. 

Since the orthography varies across languages and is usually not an accurate reflection of pronunciation, many low-resource languages are not reaping the full benefits of large-scale multilingual data in this cross-modal task in text-based models. Close inspection shows that the text-based model generalizes well to Hausa (Latin alphabet) but significantly underperforms in languages with non-Latin alphabet, such as Tamil, Vietnamese, Japanese, Arabic, and Cantonese. 

In contrast, the phoneme-based model achieves near-perfect performance in retrieval in almost all seen and unseen languages (see Figure~\ref{fig:hours}), making them extremely useful in low-resource and zero-resource scenarios. The efficiency of IPA representations in multilingual settings has also been observed in ASR \cite{feng23_interspeech}.

\paragraph{Does multilingual models always hold advantages over monolingual models?} At least in the current study, multilingual models might not hold an apparent advantage over well-engineered monolingual models in high-resource languages. As shown in Table~\ref{tab:libriphrase} and Table~\ref{tab:timit}, compared to other state-of-the-art KWS and forced alignment models,  \ourclap{} and \ouraligner{} was not able to outperform well-engineered monolingual models. Our multilingual models have not been trained on the training set of LibriPhrase or TIMIT, so some of the performance gaps might be caused by domain mismatch. Even with zero adaptations, multilingual models achieve close performance to monolingual models, suggesting that our approach is promising and may reach better results if scaled up.

\paragraph{Should we scale up the number of languages or number of training hours?} We compared \ourclap{} only trained on VoxCommunis \cite{ahn-chodroff-2022-voxcommunis-2} or \fleurs{}. VoxCommunis has almost twice as many hours as \fleurs{} with roughly half of the languages. In Table~\ref{tab:libriphrase} and~\ref{tab:unseen_eval}, \ourclap{}-\textsc{Vc} trained on more hours of speech generally has similar performance as \ourclap{}-\textsc{Fleurs} trained on a subset of the \ipapack{} across metrics, which suggests that creating high-quality data is effective in achieving good performance. But this finding also suggests that we can achieve good crosslinguistic generalizability with fewer languages but longer hours using phoneme modeling. Given the empirical data distributions in real-life settings, scaling up training hours in a dozen of languages is much easier than scaling up the number of languages. The practical implication is that we might be able to build multilingual speech processing systems for many low-resource or zero-resource languages with large-scale data in a dozen relatively high-resource languages. 

\paragraph{Is it feasible to scale up the creation of good-quality phonemic transcriptions in world languages?} Despite our attempt, there are still multiple challenges for creating phonemic transcriptions. During our dataset construction, we were unable to process many languages due to the lack of pronunciation dictionaries, text transcriptions, or relevant NLP tools, especially the lack of good word segmentation tools for some East/Southeast Asian languages like Khmer. While available large-scale speech corpora nowadays encompass more than 1000 languages \cite{salesky-etal-2020-corpus,pratap2023scaling}, textual or phonemic labels cannot be easily obtained for most of them, limiting their usage in many research applications. 

Even for high-resource languages, preprocessing multilingual texts and normalizing the Unicode encodings for IPA symbols usually take tremendous effort, not to mention verifying these phonemic transcriptions for audio recordings. %The automatic G2P conversion only makes use of textual data and its predictions can deviate from the actual pronunciations in continuous speech. 
It remains unclear how biases or noises in G2P predictions will propagate to downstream multilingual tasks. Our endeavor marks a small step in creating good-quality phonemic transcriptions for more languages. However, there is still much work to be done to include a broader array of languages worldwide and to improve the quality of transcriptions. 

\section{Conclusions}
With the carefully curated \ipapack{}, we show that using IPA symbols as modeling units can effectively enable \ourclap{} and \ouraligner{} to generalize to unseen languages, highlighting the benefits of incorporating linguistic knowledge into deep learning methods. We believe that the \ipapack{} has great potential to benefit more tasks in multilingual speech processing, such as multilingual phoneme recognition, speech synthesis, and documenting endangered languages. In the future, we will continue to expand our dataset and models to include more diverse languages.

\section{Ethical statement}
\paragraph{Data Governance} We adhered strictly to ethical practices in curating our datasets. The original FLEURS \cite{conneau2023FLEURS}, MSWC \cite{mazumder2021multilingual}, DoReCo \cite{paschen2020building} and VoxCommunis \cite{ahn-chodroff-2022-voxcommunis-2} corpora are distributed under the Creative Commons licenses. Therefore, we are permitted to re-process and re-distribute the original dataset with proper attributions. Some languages in the DoReCo corpus are under a Creative Commons Non-Commercial license. We reserved these languages to the test set in our corpora, such that our models have not been trained on data under commercially restrictive licenses. As required, we have also cited every individual language from the DoReCo Corpus in Table~\ref{app:doreco_stats}. 

\paragraph{Potential Impact} We believe that our dataset and models will contribute to the endeavor of building fair and inclusive speech processing systems for all languages and facilitating the documentation of endangered languages. However, we are aware that multilingual keyword-spotting technology could potentially be misused as surveillance tools for monitoring speech recordings in more languages, posing risks to users. 

\section{Limitations}
Our study is still limited in several aspects. First, while we tried our best to inspect a subset of our dataset, it was impossible for us to examine all datasets in great detail. As a result, the constructed dataset might still be flawed in terms of audio quality and transcription quality (and many unicode errors). Secondly, the proposed models are still not optimized in terms of computational efficiency. Since most KWS applications are running on mobile devices with limited computational power, the proposed models still have too many model parameters to run efficiently on mobile devices. Moreover, speech sequences are usually much longer than text sequences. Self-attention with quadratic complexity might not be the most suitable architecture for processing speech. More efforts are needed to make such multilingual models efficient.

Thirdly, the number of languages studied in our paper is still limited and might be biased towards languages that are relatively high-resource. They are not representative of the global language landscape. There are many more low-resource or endangered languages we are not able to include due to the lack of various resources. To promote linguistic inclusion and fairness, we will continue to improve the language diversity of our research in the future.

\section*{Acknowledgements}
This research was enabled in part through the computational resources and services provided by Advanced Research Computing at the University of British Columbia and in part through the support provided by the Digital Research Alliance of Canada. This study also benefits from the cloud computing credits awarded by Microsoft Azure through a pilot program with UBC Advanced Research Computing. 

The authors would like to thank the four anonymous reviewers as well as the area chairs for their thoughtful comments and discussions, which helped improve this article considerably. We thank Emily P. Ahn and Eleanor Chodroff for creating and releasing the VoxCommunis, which inspired our data creation process. Finally, we acknowledge that this work is impossible without the pioneering efforts of many language researchers who collected and shared speech corpora across world languages. 

\bibliography{anthology,custom}
\bibliographystyle{acl_natbib}

\newpage
\appendix

\section{Dataset statistics}
\label{app:dataset_stats}

Table~\ref{app:mswc_stats}, ~\ref{app:fleurs_stats} ,~\ref{app:wds_stats}, and ~\ref{app:doreco_stats} provide tabulated summaries of the detailed statistics of our curated datasets.

\section{Training hyperparameters}
\label{app:training}
For pre-training, we trained three variants of BERT from scratch using only phonemic transcriptions. We adopted the AdamW optimizer with an initialized learning rate of $1e-4$ and cosine scheduling with a warm-up step of 1000. All models were trained for 60k iterations before stopping. All training processes were completed on a single V100 GPU of 32 GB. 

All hyperparameters for \ourclap{} models were listed in Table~\ref{tab:hyper}. By default, all models were trained on a single V100 with 32GB of memory. The training time for \ourclap{} in 100k steps ranged from 17 hours for \ourclap{}-tiny to 41 hours for \ourclap{}-small. 

All hyperparameters for \ouraligner{} models were listed in Table~\ref{tab:hyper_align}. All models were trained on a single V100 with 32GB of memory. The training time for \ouraligner{} before early stopping ranged from 5 hours for \ourclap{}-tiny to 12 hours for \ourclap{}-small.

\begin{table*}[]
\begin{tabular}{llll}\toprule
            Hyperparameters                             & \ourclap{}-tiny & \ourclap{}-base & \ourclap{}-small \\\midrule
Hidden dimensions                        & 384           & 512           & 768            \\
Num. Layers                              & 4             & 6             & 12             \\
Num. Att. Heads                          & 6             & 8             & 12             \\
Intermediate size                        & 1536          & 2048          & 3072           \\
Parameters & 16M & 28.5M & 96.2M \\
Initial learning rate                    & \multicolumn{3}{l}{1e-4}                       \\
Scheduler                                & \multicolumn{3}{l}{Cosine Scheduler}           \\
Warm-up steps                            & \multicolumn{3}{l}{500}                        \\
Total training steps                     & \multicolumn{3}{l}{100k}                       \\
\fleurs{} batch size                      & 64 & 64             & 32            \\
\mswc{} batch size                        & 512 & 512          &256               \\
\doreco{} batch size                      & 64 & 64 & 32                         \\
VoxCommunis batch size                      & 64 & 64 & 32                         \\
Gradient checkpointing                   & \multicolumn{3}{l}{True}                       \\
Mixed Precision                          & \multicolumn{3}{l}{float16}                    \\
Max. Gradient Norm for Gradient Clipping & \multicolumn{3}{l}{10}    \\\bottomrule                    
\end{tabular}
\caption{Hyperparameters for training \ourclap{} models.}
\label{tab:hyper}
\end{table*}

\begin{table*}[]
\begin{tabular}{llll}\toprule
            Hyperparameters                             & \ourclap{}-tiny & \ourclap{}-base & \ourclap{}-small \\\midrule
Hidden dimensions                        & 384           & 512           & 768            \\
Num. Layers                              & 4             & 6             & 12             \\
Num. Att. Heads                          & 6             & 8             & 12             \\
Intermediate size                        & 1536          & 2048          & 3072           \\
Parameters & 16M & 28.5M & 96.2M \\
Initial learning rate                    & \multicolumn{3}{l}{1e-5}                       \\
Scheduler                                & \multicolumn{3}{l}{Cosine Scheduler}           \\
Warm-up steps                            & \multicolumn{3}{l}{100}                        \\
Maximum training steps                     & \multicolumn{3}{l}{10k}                       \\
batch size                      & 128 \\         
Gradient checkpointing                   & \multicolumn{3}{l}{True}                       \\
Mixed Precision                          & \multicolumn{3}{l}{float16}                    \\
Max. Gradient Norm for Gradient Clipping & \multicolumn{3}{l}{10}    \\
Early stopping & True \\
Stopping Criteria & \multicolumn{3}{l}{the highest F1 on the TIMIT training set} \\\bottomrule                    
\end{tabular}
\caption{Hyperparameters for training \ouraligner{} models.}
\label{tab:hyper_align}
\end{table*}

\onecolumn
{\centering
{\small
\begin{table*}[]
\begin{adjustbox}{width=0.99\textwidth,center}
\begin{tabular}{llllllll}
\toprule
Language         & ISO 639-3 & Family        & Train (hrs)  & Dev (hrs)  & Test (hrs)  & Avg.Phones        & Avg.Dur. (s)  \\\midrule
Arabic           & ara       & Indo-European & 0.79   & 0.11  & 0.11  & 6.57 (1.58) & 1 (0) \\
Catalan          & cat       & Indo-European & 61.38  & 8.12  & 8.07  & 7.51 (2.34) & 1 (0) \\
Czech            & cze       & Indo-European & 3.01   & 0.41  & 0.39  & 6.2 (1.92)  & 1 (0) \\
Dutch            & dut       & Indo-European & 6.42   & 0.84  & 0.84  & 7.2 (2.79)  & 1 (0) \\
English          & eng       & Indo-European & 125.61 & 16.29 & 16.52 & 6.84 (2.34) & 1 (0) \\
Esperanto        & epo       & Constructed   & 8.48   & 1.13  & 1.12  & 6.95 (1.97) & 1 (0) \\
Estonian         & est       & Uralic        & 2.51   & 0.34  & 0.33  & 6.35 (2.13) & 1 (0) \\
French           & fra       & Indo-European & 62.69  & 8.31  & 8.31  & 5.8 (2.03)  & 1 (0) \\
German           & ger       & Indo-European & 83.23  & 10.99 & 10.96 & 8.81 (3.41) & 1 (0) \\
Irish            & gle       & Indo-European & 0.48   & 0.07  & 0.07  & 4.17 (1.45) & 1 (0) \\
Greek            & gre       & Indo-European & 0.71   & 0.1   & 0.1   & 5.71 (2.09) & 1 (0) \\
Interlingua      & ina       & Constructed   & 0.53   & 0.06  & 0.05  & 5.8 (1.84)  & 1 (0) \\
Indonesian       & ind       & Austronesian  & 1.74   & 0.25  & 0.25  & 6.04 (1.88) & 1 (0) \\
Italian          & ita       & Indo-European & 18.42  & 2.46  & 2.43  & 7.35 (2.45) & 1 (0) \\
Kyrgyz           & kir       & Turkic        & 1.52   & 0.23  & 0.21  & 6.89 (2.43) & 1 (0) \\
Lithuanian       & lit       & Indo-European & 0.7    & 0.04  & 0.06  & 6.88 (2.22) & 1 (0) \\
Maltese          & mlt       & Afro-Asiatic  & 1.12   & 0.15  & 0.16  & 6.04 (2.75) & 1 (0) \\
Mongolian        & mon       & Mongolic      & 1.48   & 0.2   & 0.21  & 5.5 (1.73)  & 1 (0) \\
Polish           & pol       & Indo-European & 14.39  & 1.93  & 1.94  & 6.9 (2.09)  & 1 (0) \\
Portuguese       & por       & Indo-European & 7.16   & 0.95  & 0.95  & 6.52 (2.07) & 1 (0) \\
Romanian         & ron       & Indo-European & 0.5    & 0.1   & 0.08  & 6.43 (2.31) & 1 (0) \\
Russian          & rus       & Indo-European & 18.48  & 2.46  & 2.44  & 8.48 (2.88) & 1 (0) \\
Slovak           & slo       & Indo-European & 0.08   & 0.01  & 0.01  & 6.28 (2.14) & 1 (0) \\
Slovenian        & slv       & Indo-European & 0.27   & 0.05  & 0.05  & 5.16 (1.56) & 1 (0) \\
Spanish          & spa       & Indo-European & 40.04  & 5.35  & 5.32  & 7.64 (2.49) & 1 (0) \\
Swedish          & swe       & Indo-European & 1.18   & 0.16  & 0.16  & 5.36 (1.93) & 1 (0) \\
Tatar            & tat       & Turkic        & 3.76   & 0.5   & 0.48  & 6.18 (1.84) & 1 (0) \\
Turkish          & tur       & Turkic        & 2.82   & 0.38  & 0.39  & 6.95 (2.27) & 1 (0) \\
Ukrainian        & ukr       & Indo-European & 1.87   & 0.25  & 0.26  & 6.76 (2.3)  & 1 (0) \\
Welsh            & wel       & Indo-European & 13.6   & 1.8   & 1.81  & 5.76 (1.96) & 1 (0) \\
Mandarin & cmn       & Sino-Tibetan  & 0.4    & 0.05  & 0.05  & 8.68 (2.26) & 1 (0) \\\bottomrule
\end{tabular}
\end{adjustbox}
\caption{Statistics of languages in \mswc{}. All samples are padded to be clips of 1 second. (Avg.Phones: average number of phonemes in each word; Avg.Dur.: average duration of each clip). }
\label{app:mswc_stats}
\end{table*}
}}
\onecolumn
{\centering
{\small
%\begin{table*}[]

\begin{longtable}{llllll}
%\begin{adjustbox}{width=0.99\textwidth,center}
\toprule
Language & ISO 639-3  & Family & Train (hrs)  &  Avg. Dur (s) & Avg. Phones \\\midrule
Abkhaz                  & abk & Northwest Caucasian & 0.62   & 7.33 & 51.93 \\
Bashkir                 & bak & Turkic              & 137.79 & 4.35 & 35.78 \\
Belarusian              & bel & Indo-European       & 132.21 & 5.48 & 49.15 \\
Bulgarian               & bul & Indo-European       & 3.5    & 5.05 & 47.74 \\
Catalan                 & cat & Indo-European       & 2.08   & 5.39 & 44.35 \\
Czech                   & ces & Indo-European       & 16.51  & 4.75 & 44    \\
Chuvash                 & chv & Turkic              & 0.37   & 4.2  & 36.97 \\
Greek                   & ell & Indo-European       & 1.57   & 3.99 & 29.13 \\
Basque                  & eus & Language isolate    & 12.66  & 5.2  & 47.36 \\
Guarani                 & grn & Tupian              & 1.81   & 3.97 & 26.91 \\
Hausa                   & hau & Afro-Asiatic        & 1.71   & 4.27 & 32.1  \\
Hindi                   & hin & Indo-European       & 2.73   & 3.75 & 33.69 \\
Sorbian (Upper Sorbian) & hsb & Indo-European       & 1.48   & 6.61 & 55.01 \\
Hungarian               & hun & Uralic              & 25.06  & 4.76 & 37.62 \\
Indonesian              & ind & Austronesian        & 5.09   & 5.69 & 53.31 \\
Italian                 & ita & Indo-European       & 192.69 & 5.24 & 49.13 \\
Georgian                & kat & Kartvelian          & 1.62   & 5.77 & 53.7  \\
Kazakh                  & kaz & Turkic              & 0.29   & 4.93 & 33.95 \\
Kurmanji (Kurdish)      & kmr & Indo-European       & 2.83   & 4.47 & 28.16 \\
Kyrgyz                  & kir & Turkic              & 2.25   & 4.67 & 43.42 \\
Marathi                 & mar & Indo-European       & 3.66   & 5.97 & 52.99 \\
Maltese                 & ml  & Afro-Asiatic        & 2.41   & 4.48 & 36.88 \\
Erzya                   & myv & Uralic              & 1.97   & 5.73 & 46.41 \\
Dutch                   & nld & Indo-European       & 34.94  & 4.4  & 47.47 \\
Punjabi                 & pan & Indo-European       & 0.96   & 5.29 & 26.53 \\
Polish                  & pol & Indo-European       & 14.26  & 5.21 & 46.58 \\
Portuguese              & por & Indo-European       & 12.31  & 4.33 & 32.45 \\
Romanian                & ron & Indo-European       & 4.99   & 4.01 & 35.98 \\
Russian                 & rus & Indo-European       & 24.41  & 5.44 & 56.61 \\
Swedish                 & swe & Indo-European       & 5.84   & 3.84 & 31.24 \\
Swahili                 & swa & Niger-Congo         & 52.8   & 5.44 & 47.43 \\
Tamil                   & tam & Dravidian           & 61.39  & 6.57 & 55.43 \\
Thai                    & tha & Kra-Dai             & 16.71  & 3.91 & 26.58 \\
Turkish                 & tur & Turkic              & 0.98   & 3.19 & 30.43 \\
Tatar                   & tat & Turkic              & 10.09  & 3.8  & 31.5  \\
Uyghur                  & uig & Turkic              & 2.43   & 5.85 & 49.21 \\
Ukrainian               & ukr & Indo-European       & 4.22   & 4.67 & 39.54 \\
Vietnamese              & vie & Austroasiatic       & 4.6    & 4.53 & 25.54
  \\\bottomrule     
%\end{tabular}

\caption{Detailed statistics of a selected subset of VoxCommunis \cite{ahn-chodroff-2022-voxcommunis-2}.  (Avg.Phones: average number of phonemes in each word; Avg.Dur.: average duration of each clip).}
\label{app:wds_stats}
%\end{adjustbox}
\end{longtable}

}}
\onecolumn
{\centering
{\footnotesize
\begin{longtable}{p{20mm}p{10mm}p{10mm}p{15mm}p{15mm}p{20mm}lp{20mm}}
\toprule
Language & ISO 693-3 & Avg. Dur (s) & Total duration (hrs) & Avg. Phones & Family & Split & Citation \\
\midrule
Komnzo & tci & 2.59 & 0.27 & 29.99 & Yam & train & \citep{doreco-komn1238} \\
Vera'a & vra & 3.55 & 0.57 & 43.03 & Austronesian & train & \citep{doreco-vera1241} \\
Sanzhi Dargwa & na & 4.85 & 0.17 & 44.82 & Nakh-Daghestanian & train & \citep{doreco-sanz1248} \\
Urum & uum & 4.63 & 0.37 & 45.75 & Turkic & test & \citep{doreco-urum1249} \\
Beja & bej & 2.32 & 0.36 & 24.97 & Afro-Asiatic & test & \citep{doreco-beja1238} \\
Light Warlpiri & na & 3.47 & 0.47 & 32.75 & Mixed Language & train & \citep{doreco-ligh1234} \\
Kamas & xas & 3.60 & 0.84 & 24.71 & Uralic & train & \citep{doreco-kama1351} \\
Nafsan (South Efate) & erk & 6.10 & 0.36 & 50.83 & Austronesian & test & \citep{doreco-sout2856} \\
Tabasaran & tab & 4.16 & 0.21 & 42.31 & Nakh-Daghestanian & train & \citep{doreco-taba1259} \\
Savosavo & svs & 5.17 & 0.82 & 49.35 & Isolate & train & \citep{doreco-savo1255} \\
Sümi & nsm & 2.74 & 0.14 & 32.59 & Sino-Tibetan & train & \citep{doreco-sumi1235} \\
French (Swiss) & fra & 2.75 & 0.31 & 32.61 & Indo-European & test & \citep{doreco-stan1290} \\
Northern Alta & aqn & 2.78 & 1.04 & 25.94 & Austronesian & train & \citep{doreco-nort2875} \\
Jejuan & jje & 2.59 & 0.03 & 24.43 & Koreanic & train & \citep{doreco-jeju1234} \\
Jahai & jhi & 3.61 & 0.45 & 32.74 & Austroasiatic & test & \citep{doreco-jeha1242} \\
Nisvai & none & 3.11 & 0.56 & 42.22 & Austronesian & test & \citep{doreco-nisv1234} \\
Warlpiri & wbp & 3.64 & 0.94 & 30.84 & Pama-Nyungan & test & \citep{doreco-warl1254} \\
Fanbyak & fnb & 2.81 & 0.22 & 27.29 & Austronesian & train & \citep{doreco-orko1234} \\
Bora & boa & 4.40 & 0.34 & 41.49 & Boran & train & \citep{doreco-bora1263} \\
Yongning Na & nru & 4.23 & 0.30 & 33.15 & Sino-Tibetan & train & \citep{doreco-yong1270} \\
Dalabon & ngk & 2.46 & 0.08 & 23.46 & Gunwinyguan & train & \citep{doreco-ngal1292} \\
Sadu & na & 2.75 & 0.15 & 22.78 & Sino-Tibetan & train & \citep{doreco-sadu1234} \\
Teop & tio & 2.96 & 0.65 & 30.62 & Austronesian & train & \citep{doreco-teop1238} \\
Cashinahua & cbs & 3.58 & 0.73 & 33.55 & Pano-Tacanan & train & \citep{doreco-cash1254} \\
Dolgan & dlg & 4.24 & 0.69 & 43.55 & Turkic & test & \citep{doreco-dolg1241} \\
Anal & anm & 3.02 & 0.37 & 26.43 & Sino-Tibetan & train & \citep{doreco-anal1239} \\
Baïnounk Gubëeher & bab & 3.13 & 0.40 & 30.95 & Atlantic-Congo & train & \citep{doreco-bain1259} \\
Texistepec Popoluca & poq & 2.65 & 0.08 & 28.50 & Mixe-Zoque & train & \citep{doreco-texi1237} \\
Daakie & ptv & 3.22 & 0.22 & 34.87 & Austronesian & train & \citep{doreco-port1286} \\
Ning & ngh & 2.67 & 0.12 & 22.96 & Tuu & train & \citep{doreco-nngg1234} \\
Ruuli & ruc & 3.13 & 0.32 & 34.99 & Atlantic-Congo & train & \citep{doreco-ruul1235} \\
Cabécar & cjp & 3.61 & 0.38 & 39.62 & Chibchan & test & \citep{doreco-cabe1245} \\
Evenki & evn & 3.89 & 0.66 & 31.71 & Tungusic & train & \citep{doreco-even1259} \\
Arapaho & arp & 3.99 & 0.87 & 32.95 & Algic & train & \citep{doreco-arap1274} \\
Svan & sva & 4.77 & 0.56 & 47.85 & Kartvelian & train & \citep{doreco-svan1243} \\
Resígaro & rgr & 5.45 & 1.27 & 33.31 & Arawakan & train & \citep{doreco-resi1247} \\
Yali (Apahapsili) & na & 2.38 & 0.04 & 32.34 & Nuclear Trans New Guinea & test & \citep{doreco-apah1238} \\
Asimjeeg Datooga & na & 2.81 & 0.28 & 28.30 & Nilotic & train & \citep{doreco-tsim1256} \\
Northern Kurdish (Kurmanji) & kmr & 4.39 & 0.54 & 50.76 & Indo-European & test & \citep{doreco-nort2641} \\
Gorwaa & gow & 2.95 & 0.28 & 31.29 & Afro-Asiatic & train & \citep{doreco-goro1270} \\
Pnar & pbv & 8.28 & 0.29 & 72.74 & Austroasiatic & test & \citep{doreco-pnar1238} \\
Kakabe & kke & 4.14 & 0.59 & 33.07 & Mande & train & \citep{doreco-kaka1265} \\
Mojeño Trinitario & trn & 5.66 & 0.65 & 48.42 & Arawakan & train & \citep{doreco-trin1278} \\
\bottomrule
\caption{Detailed statistics of \textsc{Doreco-Ipa}.  (Avg. Phones: average
number of phonemes in each word; Avg.Dur.: average duration of each clip).}
\label{app:doreco_stats}
\end{longtable}
}}

\onecolumn
{\centering
{\footnotesize

%\begin{table*}[]
%\begin{adjustbox}{width=0.99\textwidth,center}
\begin{longtable}{lllllllll}
\toprule
Language & ISO 639-3  & Family & Train (hrs) & Dev (hrs) & Test (hrs) &  Avg. Dur (s) & Avg. Phones \\\midrule
Afrikaans         & afr & Indo-European & 2.71  & 0.48 & 0.66 & 11.95 & 4.69  \\
Amharic           & amh & Afro-Asiatic  & 8.26  & 0.57 & 1.28 & 11.91 & 7.25  \\
Arabic            & ara & Afro-Asiatic  & 4.93  & 0.75 & 1.12 & 10.2  & 6.63  \\
Azerbaijani       & aze & Turkic        & 6.89  & 1.1  & 2.42 & 12.27 & 6.33  \\
Belarusian        & bel & Indo-European & 7.3   & 1.37 & 3.11 & 13.87 & 6.08  \\
Bulgarian         & bul & Indo-European & 7.05  & 0.85 & 1.44 & 10.65 & 5.47  \\
Bengali           & ben & Indo-European & 8.18  & 1.21 & 2.75 & 12.67 & 5.84  \\
Bosnian           & bos & Indo-European & 7.57  & 1.1  & 2.47 & 11.4  & 5.42  \\
Catalan           & cat & Indo-European & 5.77  & 1.09 & 2.43 & 11.34 & 4.64  \\
Cebuano           & ceb & Austronesian  & 9.33  & 0.72 & 1.77 & 13.26 & 4.54  \\
Mandarin Chinese  & cmn & Sino-Tibetan  & 6.04  & 0.87 & 2    & 10.37 & 3.81  \\
Czech             & cze & Indo-European & 6.38  & 0.82 & 1.91 & 10.76 & 5.58  \\
Welsh             & wel & Indo-European & 9.12  & 1.49 & 3.32 & 12.98 & 4.27  \\
Danish            & dan & Indo-European & 5.75  & 0.99 & 2.26 & 10.69 & 4.4   \\
German            & ger & Indo-European & 6.88  & 1.06 & 2.46 & 11.16 & 5.61  \\
Greek             & gre & Indo-European & 7.51  & 0.64 & 1.47 & 10.69 & 5.14  \\
English           & eng & Indo-European & 5.64  & 0.88 & 1.39 & 9.79  & 4.41  \\
Spanish           & spa & Indo-European & 6.73  & 1.17 & 2.45 & 11.24 & 4.87  \\
Estonian          & est & Uralic        & 5.38  & 1.02 & 2.37 & 10.57 & 6.35  \\
Fula              & ful & Niger-Congo   & 10.27 & 0.84 & 2.12 & 14.35 & 4.16  \\
Finnish           & fin & Uralic        & 6.75  & 1.18 & 2.58 & 11.61 & 7.04  \\
Irish             & gle & Indo-European & 9.31  & 1.24 & 2.76 & 14.54 & 3.68  \\
Galician          & glg & Indo-European & 5.12  & 0.89 & 2.06 & 10.31 & 5.01  \\
Hausa             & hau & Afro-Asiatic  & 10.09 & 1.25 & 2.47 & 15.21 & 4.36  \\
Hindi             & hin & Indo-European & 5.14  & 0.63 & 1.11 & 11.01 & 4.08  \\
Croatian          & hrv & Indo-European & 8.78  & 0.85 & 1.98 & 11.14 & 5.43  \\
Hungarian         & hun & Uralic        & 7.01  & 1.14 & 2.45 & 10.85 & 5.76  \\
Indonesian        & ind & Austronesian  & 6.94  & 0.97 & 1.89 & 12.18 & 5.79  \\
Icelandic         & ice & Indo-European & 2.11  & 0.1  & 0.14 & 10.8  & 5.53  \\
Italian           & ita & Indo-European & 6.86  & 1.31 & 2.8  & 11.52 & 4.97  \\
Japanese          & jpn & Japonic       & 5.06  & 0.67 & 1.52 & 11.63 & 3.63  \\
Javanese          & jav & Austronesian  & 8.6   & 0.94 & 2.22 & 12.98 & 5.47  \\
Georgian          & geo & Kartvelian    & 3.87  & 0.99 & 2.37 & 11.31 & 7.14  \\
Kazakh            & kaz & Turkic        & 8.91  & 1.29 & 3.02 & 13.55 & 6.78  \\
Korean            & kor & Koreanic      & 5.68  & 0.57 & 1.03 & 12.14 & 7.17  \\
Kyrgyz            & kir & Turkic        & 6.99  & 1.1  & 2.52 & 11.45 & 6.83  \\
Lao               & lao & Kra-Dai       & 5.58  & 0.47 & 1.09 & 13.41 & 21.33 \\
Lithuanian        & lit & Indo-European & 7.28  & 0.97 & 2.32 & 10.96 & 6.33  \\
Maori             & mri & Austronesian  & 13.34 & 1.86 & 4.53 & 19.34 & 3.48  \\
Macedonian        & mac & Indo-European & 5.14  & 1.05 & 2.47 & 10.5  & 5.35  \\
Malayalam         & mal & Indo-European & 7.37  & 1.36 & 2.86 & 12.28 & 10.23 \\
Mongolian         & mon & Mongolic      & 8.63  & 0.97 & 2.21 & 12.19 & 5.52  \\
Marathi           & mar & Indo-European & 9.48  & 1.23 & 3.04 & 12.96 & 6     \\
Malay             & msa & Austronesian  & 7.28  & 0.79 & 1.82 & 11.8  & 5.99  \\
Maltese           & mlt & Afro-Asiatic  & 7.5   & 1.24 & 2.81 & 12.31 & 4.68  \\
Burmese           & bur & Sino-Tibetan  & 10.07 & 1.49 & 3.25 & 14.56 & 12.17 \\
Norwegian         & nob & Indo-European & 7.96  & 0.43 & 0.93 & 12.06 & 4.54  \\
Dutch             & dut & Indo-European & 5.81  & 0.38 & 0.77 & 9.18  & 4.89  \\
Nyanja            & nya & Niger-Congo   & 8.23  & 1.2  & 2.77 & 14.53 & 5.99  \\
Oromo             & orm & Afro-Asiatic  & 5.11  & 0.05 & 0.13 & 13.46 & 5.41  \\
Oriya             & ori & Indo-European & 2.42  & 1    & 2.25 & 11.33 & 6.5   \\
Punjabi           & pan & Indo-European & 4.96  & 0.63 & 1.48 & 11.49 & 4.07  \\
Polish            & pol & Indo-European & 7.23  & 0.73 & 1.63 & 10.71 & 5.66  \\
Portuguese        & por & Indo-European & 7.77  & 1.06 & 2.5  & 12.49 & 4.7   \\
Romanian          & ron & Indo-European & 7.65  & 0.88 & 1.95 & 11.46 & 5.31  \\
Russian           & rus & Indo-European & 6.28  & 0.92 & 1.94 & 10.97 & 6.24  \\
Sindhi            & snd & Indo-European & 9.15  & 1.1  & 2.55 & 12.15 & 4.41  \\
Slovak            & slo & Indo-European & 4.55  & 0.92 & 2.1  & 10.8  & 5.58  \\
Slovenian         & slv & Indo-European & 5.78  & 0.74 & 1.78 & 10.2  & 5.43  \\
Shona             & sna & Niger-Congo   & 7.56  & 1.27 & 3.03 & 14.12 & 6.88  \\
Somali            & som & Afro-Asiatic  & 9.84  & 1.26 & 3.03 & 14.04 & 4.77  \\
Serbian           & srp & Indo-European & 8.14  & 0.7  & 1.66 & 12.05 & 5.25  \\
Swedish           & swe & Indo-European & 6.34  & 0.79 & 1.82 & 11.64 & 5.08  \\
Swahili           & swa & Niger-Congo   & 10.1  & 0.69 & 1.54 & 14.72 & 5.15  \\
Tamil             & tam & Indo-European & 6.34  & 1.04 & 1.61 & 12.5  & 8.12  \\
Telugu            & tel & Indo-European & 5.87  & 0.75 & 1.11 & 11.64 & 7.03  \\
Tajik             & tgk & Indo-European & 6.52  & 0.77 & 1.96 & 13.43 & 5.39  \\
Thai              & tha & Kra-Dai       & 6.21  & 1.14 & 2.56 & 11.34 & 4.83  \\
Turkish           & tur & Turkic        & 6.43  & 0.94 & 2.09 & 11.77 & 6.48  \\
Ukrainian         & ukr & Indo-European & 6.7   & 0.78 & 1.78 & 10.82 & 5.87  \\
Urdu              & urd & Indo-European & 5.34  & 0.64 & 0.66 & 11.17 & 3.9   \\
Uzbek             & uzb & Turkic        & 7.6   & 0.99 & 2.25 & 11.8  & 6.58  \\
Vietnamese        & vie & Austroasiatic & 6.71  & 1.01 & 2.33 & 10.97 & 4.07  \\
Xhosa             & xho & Niger-Congo   & 9.78  & 1.27 & 2.91 & 12.96 & 7.19  \\
Yoruba            & yor & Niger-Congo   & 8.46  & 1.56 & 3.26 & 15.48 & 3.48  \\
Cantonese Chinese & yue & Sino-Tibetan  & 5.56  & 0.93 & 2.07 & 12.31 & 3.96  \\
Zulu              & zul & Niger-Congo   & 11.05 & 1.31 & 3.03 & 17.3  & 7.23 
  \\\bottomrule     
%\end{tabular}
%\end{adjustbox}
\caption{Detailed statistics of \textsc{Fleurs-Ipa}.  (Avg.Phones: average
number of phonemes in each word; Avg.Dur.: average duration of each clip).}
\label{app:fleurs_stats}
\end{longtable}

}}

\end{document}